\documentclass[conference,a4paper]{IEEEtran}
\IEEEoverridecommandlockouts
\usepackage{cite}
\usepackage{amsmath,amssymb,amsfonts}
\usepackage{algorithm}
\usepackage{multirow}
\usepackage{algorithmic}
\usepackage{graphicx}
\usepackage{textcomp}
\usepackage{xcolor}
\usepackage[bottom=2.54cm,top=1.91cm,left=1.59cm,right=1.59cm]{geometry}
\def\BibTeX{{\rm B\kern-.05em{\sc i\kern-.025em b}\kern-.08em
    T\kern-.1667em\lower.7ex\hbox{E}\kern-.125emX}}

\makeatletter
\newcommand{\linebreakand}{%
    \end{@IEEEauthorhalign}
    \hfill\mbox{}\par
    \mbox{}\hfill\begin{@IEEEauthorhalign}
}
\makeatother

\begin{document}

\title{Digital Twin System for Home Service Robot Based on Motion Simulation\\

\thanks{This work was supported in part by the National Natural Science Foundation of China under Grant 62273203, Grant U1813215, and in part by the Special Fund for the Taishan Scholars Program of Shandong Province (ts2015110005).}
}


\author{\IEEEauthorblockN{Zhengsong Jiang\IEEEauthorrefmark{1},
                          Guohui Tian\IEEEauthorrefmark{1}\IEEEauthorrefmark{3}\thanks{\IEEEauthorrefmark{3}Guohui Tian is corresponding author.},
                          Yongcheng Cui\IEEEauthorrefmark{1},
                          Tiantian Liu\IEEEauthorrefmark{1}\IEEEauthorrefmark{4}\thanks{\IEEEauthorrefmark{4}TianTian Liu is joint corresponding author.},
                          Yu Gu\IEEEauthorrefmark{1},
                          and Yifei Wang\IEEEauthorrefmark{2}}
        \IEEEauthorblockA{\IEEEauthorrefmark{1} School of Control Science and Engineering, Shandong University, Jinan, China}
        \IEEEauthorblockA{\IEEEauthorrefmark{2} School of Information, University of California, Berkeley, USA}
        \IEEEauthorblockA{Email: zs.jiang@mail.sdu.edu.cn, g.h.tian@sdu.edu.cn, cuiyc@sdu.edu.cn\\liutiantian@sdu.edu.cn, y.gu@mail.sdu.edu.cn, Sarahwang688@berkeley.edu}
}

\maketitle

\begin{abstract}
In order to improve the task execution capability of home service robot, and to cope with the problem that purely physical robot platforms cannot sense the environment and make decisions online, a method for building digital twin system for home service robot based on motion simulation is proposed. A reliable mapping of the home service robot and its working environment from physical space to digital space is achieved in three dimensions: geometric, physical and functional. In this system, a digital space-oriented URDF file parser is designed and implemented for the automatic construction of the robot geometric model. Next, the physical model is constructed from the kinematic equations of the robot and an improved particle swarm optimization algorithm is proposed for the inverse kinematic solution. In addition, to adapt to the home environment, functional attributes are used to describe household objects, thus improving the semantic description of the digital space for the real home environment. Finally, through geometric model consistency verification, physical model validity verification and virtual-reality consistency verification, it shows that the digital twin system designed in this paper can construct the robot geometric model accurately and completely, complete the operation of household objects successfully, and the digital twin system is effective and practical.
\end{abstract}


\section{Introduction}

Home service robots, as an important medium to improve the quality of human life, are able to replace humans to complete domestic work. People not only demand that they can perform simple tasks such as sweeping floors or escorting, but also expect them to perform complex tasks such as delivering objects or preparing meals. However, in the face of increasingly complex domestic service tasks, relying solely on physical robotic platforms, the execution of tasks is highly unstable, and often unpredictable problems occur, which are likely to cause irreversible damage to expensive physical robots or the home environment, with a high degree of risk and uncertainty. Therefore, it is very necessary to design a digital twin system for home service robots to simulate various situations that may occur in real environments, to try to discover and solve problems that may occur when the physical robot platform actually operates, and to guide the physical robot to perform home service tasks reliably and efficiently.

In this paper, we argue that the main things that a robot can rely on to complete domestic service tasks are the movement of its robotic arm and the movement of its chassis. Most simulation platforms or physical engines already integrate path planning internally to enable chassis movement, so this paper focuses on how to simulate the real motion of the robotic arm in a virtual environment and build the digital twin system based on it.

Digital twin refers to the construction of a virtual mapping of the physical entity in the whole life cycle of a product or system, through data fusion, information interaction, and virtual simulation. The two are synchronized and related to achieve the integration of models, data and technology \cite{taofei_chejian}. The digital twin was originally known as the digital mirror space, which creates an accurately mapped digital space for the physical space, to describe the operational state of the physical space throughout its full life cycle \cite{dt_begin}, and was later refined by NASA in 2012 \cite{nasa}.

In the following years, a part of the research focused on the underlying theory and integrated modeling of the digital twin \cite{dt_sota}. The first generic framework for the digital twin was modeled in terms of physical entities, virtual models and connections \cite{dt3}. For full life cycle monitoring of complex objects, the five-dimensional model that adds services and digital twin data to the three-dimensional model was proposed \cite{dt5}. Among the five-dimensional models, the digital twin model is the prerequisite for realizing the digital twin on the ground, so the construction principles of the model and the construction theory are proposed \cite{dt_model}. For the complex digital twin system, \cite{complex_model} proposed to model it by decomposing it into several simple models and then fusion them. In addition, targeted modeling methods are proposed for different models in different fields \cite{dt_model_1} \cite{dt_model_2}.

Driven by the development strategies of various countries \cite{deguo4} \cite{gongye4}, the ground application of digital twin technology has become a hot spot for research. For example, digital twin technology is applied in the fields of work shop management \cite{taofei_chejian} \cite{chejian2}, power system \cite{power_system}, advanced nuclear energy \cite{heneng}, and aerospace \cite{hangkong}, to promote the improvement of intelligence in various industries. In the field of robotics, digital twin technology plays an equally large role \cite{robot0}. To improve the efficiency of design, construction and control of human-robot collaboration, digital twin system for human-robot collaboration and assembly work are constructed that can be kept up to date by continuously mapping the physical system throughout its full life cycle for rapid and continuous improvement \cite{robot1}. Through virtual reality technology, \cite{robot2} proposes a digital twin-based programming method for industrial robot demonstrations, which improves the human-robot interaction of robot demonstrations. Reference \cite{robot3} proposes a digital twin-based approach for flexible robot work cell development, which speeds up the overall commissioning or reconfiguration process by improving the work cell in digital space.

Digital twin is rapidly developing in the field of industrial robots, but it is still not targeted for home service robots. The home environment, unlike the industrial environment, is unstructured, dynamic, and has an uncertain and large number of manipulable objects, so the digital twin system construction method for industrial robots cannot be transferred to home service robots directly. In order to cope with the many characteristics of the home environment, this paper proposes the digital twin system for home service robot based on motion simulation. The system integrates multiple dimensions of information, including geometric, physical and functional. For the geometric model of the robot, a parser for URDF \cite{urdf} (Unified Robot Description Format) files is designed in this paper to build the geometric model of the robot in digital space automatically and quickly. For realizing the rational motion of the robot arm in digital space, this paper models the kinematics of the 7 Dof robot arm and proposes an improved particle swarm optimization algorithm to solve the inverse kinematics problem. In addition, to meet the characteristics of the home environment, this paper proposes to use functional attributes as the information source to describe the semantic information of household objects. Finally, the home environment is simulated in the laboratory to build the digital twin system. The validity and practicality of the system proposed in this paper are demonstrated by geometric model consistency verification, physical model validity verification and virtual-reality consistency verification.

The rest of the paper is organized as follows. The overall description of the digital system is shown in Section 2. Section 3 describe the methods for constructing home service robot digital twin system. Section 4 describes the experimental results and analysis. Finally, Section 5 describes the conclusion.

\section{Overall Description of the System}

\subsection{System Framework and Workflow}

\begin{figure*}[htbp]
    \centerline{\includegraphics[width=1\linewidth]{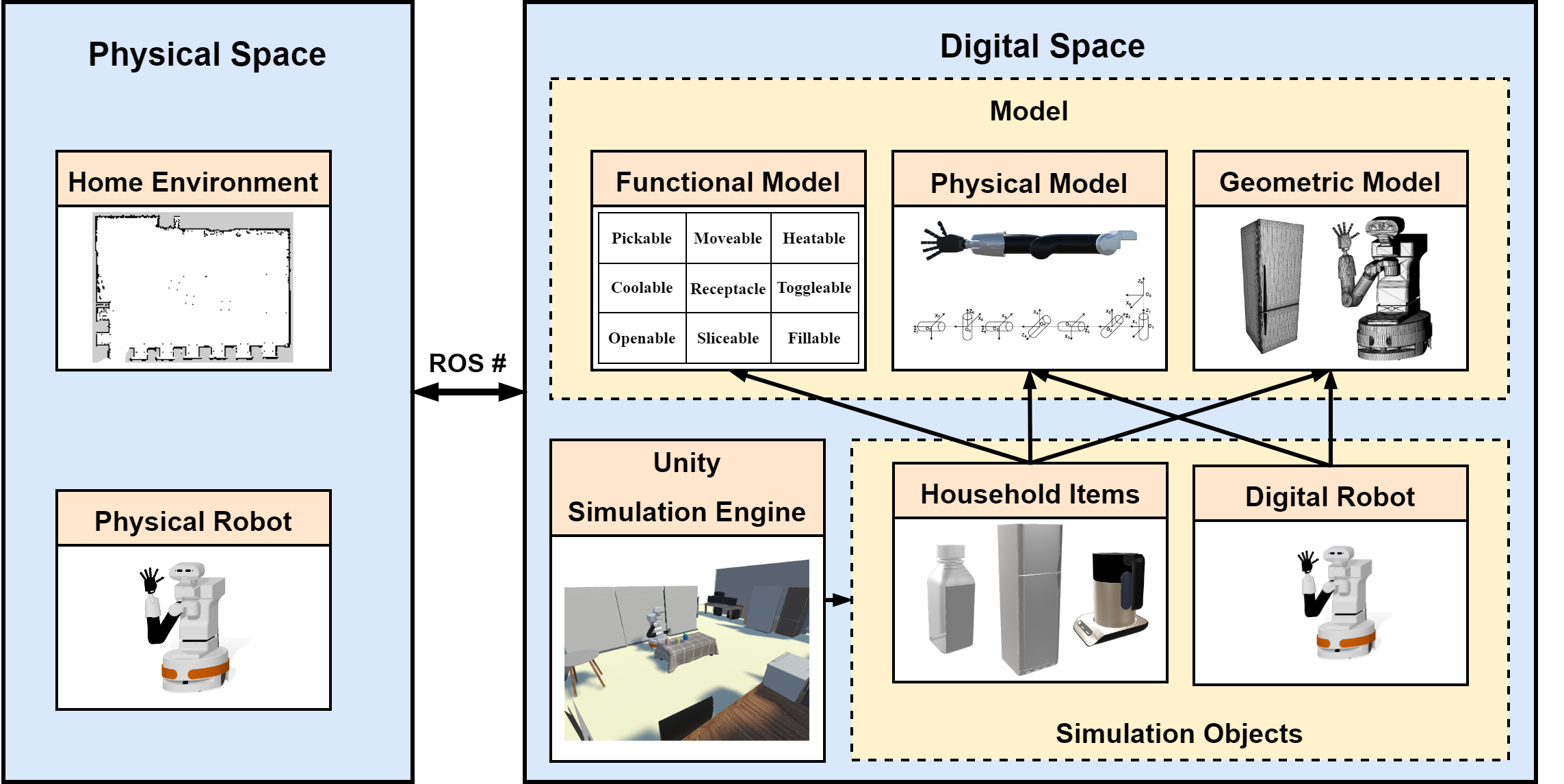}}
    \caption{Framework for the digital twin system for home service robot based on motion simulation.}
    \label{fig1}
\end{figure*}

The framework of the proposed system is shown in Fig.~\ref{fig1}, which mainly includes physical space, digital space and connections. The physical space is a dynamic system, consisting of service robots and home environment. The digital space is composed of the digital robot platform, multi-source models of household objects and the virtual home environment, which is required to map the physical space realistically in real time. The connection of this digital twin system is based on ROS\#, which reliably exchanges twin data of the two spaces.

In the construction phase of the digital twin system of the home service robot, first, geometric and physical data of the home environment and the robot platform are collected manually. Secondly, the corresponding graphical models are created by 3D modeling tools and deployed into the Unity3D-based digital space. Finally, the motion of the robotic arm is implemented in the digital space. During the operation phase of the system, the physical robotics platform monitors various operational data in the physical space in real time, including odometer information, robot arm status, and RGB-D images of the robot's viewpoint. This data is connected to the digital space via ROS\#, which helps map the physical space to the digital space with high fidelity. In the digital space, the robot simulates the operation of the real environment in the digital space to guide the physical robot platform.

\subsection{Model Composition and Correlation Analysis}

The digital twin model $M$ consists of the geometric model $G_v$, the physical model $P_v$ and the functional model $F_v$, which is an integration of the structure, properties and functions of household objects and robotic platform in physical space:

\begin{equation}
\label{equ1}
    M = G_v \cup P_v \cup F_v,
\end{equation}

where, $G_v$ is the construction of 3D models, including the shape, size, location and material properties of all objects in the physical twin space, which has the most intuitive impact on the visual effect of the digital space; $P_v$ describes the physical properties and states of the robot platform and household objects, such as gravity and collision relationships, which determine the similarity between physical and digital spaces; $F_v$ is used to describe the functional properties of household objects, such as moveable, heatable, openable, etc., which determine the behavior of the robot in digital space that is consistent with real-world common sense.

\section{Construction Method of Digital Twin System}

This section introduces the construction method of the digital twin system for home service robot, described in terms of the robot and the home environment, respectively. Among them, robot modeling contains geometry modeling and kinematic modeling. And the home environment modeling mainly contains geometric, physical and functional modeling.

\subsection{Geometric Modeling of Robot}

The robot model contains complex joint information and kinematic parameters that can be constructed with the help of URDF files. The URDF file uses XML language to define the information of Joint and Link. URDF files can be obtained from the ROS platform and can be parsed directly by the Gazebo software in the ROS system, but cannot be used directly in Unity3D. A corresponding URDF file parser, therefore, needs to be designed for the Unity3D platform.

The URDF parser uses the $System.xml$ namespace of C\# to parse the URDF file, and gets the robot's Joint and Link information, as well as the robot's description file and material file. The above information can form the basic framework of the model, so that the robot's URDF model can be imported into Unity3D as a $GameObject$.

\begin{figure}[htbp]
    \centerline{\includegraphics[width=0.5\linewidth]{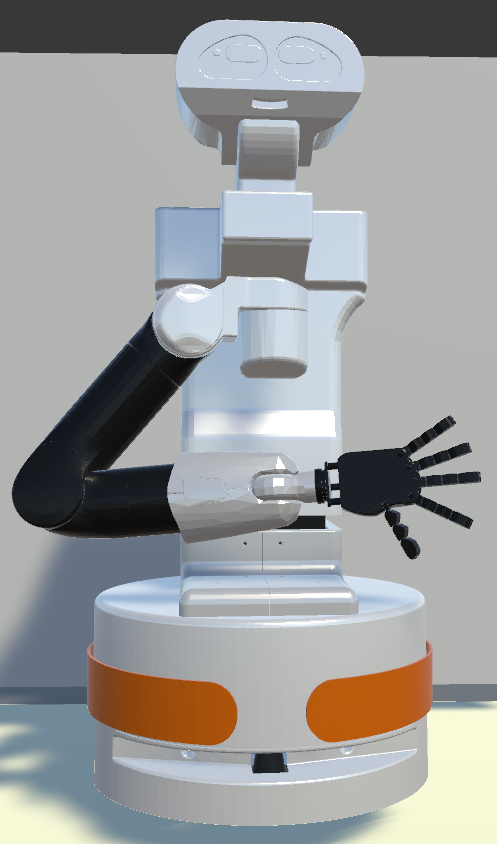}}
    \caption{Geometric model of the TIAGo robot.}
    \label{tiago}
\end{figure}

In this paper, we use the TIAGo robot as a physical robot platform that has a robotic arm with seven degrees of freedom, which excels in dynamic performance, motion planning, etc. It can grasp larger and heavier objects, and with the Hey5-type five-finger manipulator, it can also operate on tiny objects \cite{tiago}. In addition, the TIAGo robot has a PMB-2 type mobile chassis, which enables it to move flexibly in indoor environments and thus perform a variety of home service tasks. The geometric model of the TIAGo robot obtained by 3D stereoscopic display is shown in Fig.~\ref{tiago}.

\subsection{Forward Kinematic Modeling of Robot Arm}

\begin{figure}[htbp]
    \centerline{\includegraphics[width=1\linewidth]{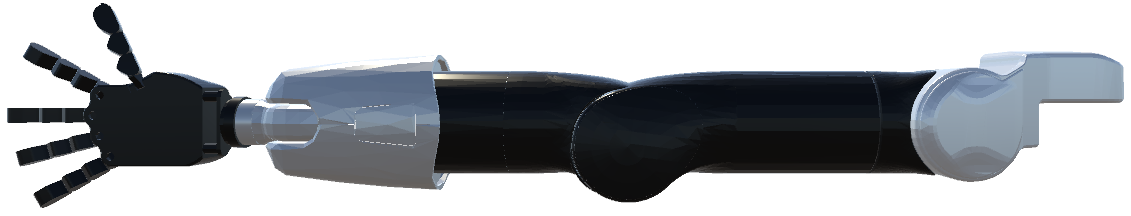}}
    \caption{TIAGo's robot arm with 7 degrees of freedom.}
    \label{tiago_arm}
\end{figure}

The robot arm of the TIAGo robot is shown in Fig.~\ref{tiago_arm} and has 7 degrees of freedom. The detailed structure can be obtained from the URDF file of the TIAGo robot.

\begin{figure}[htbp]
    \centerline{\includegraphics[width=1\linewidth]{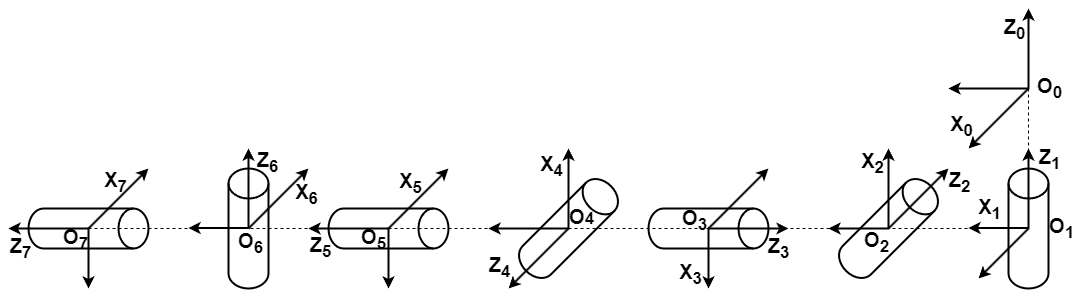}}
    \caption{Coordinate system of each joint of TIAGo robot arm.}
    \label{tiago_arm_axis}
\end{figure}

As shown in Fig.~\ref{tiago_arm_axis}, the coordinate system at each joint of the TIAGo robot arm can be established. Among them, $X_1\sim X_7$, $ Z_1\sim Z_7$ is the coordinate axes at the $1^{st}$ to $7^{th}$ joints, respectively. By denoting the 7 joint variables of the TIAGo robot arm as $\theta_1, \theta_2, ..., \theta_7$, the forward kinematic equations of this robot arm can be established using the D-H parameter method.

By reviewing the relevant information, the D-H parameter of the TIAGo robot arm are shown in Tab.~\ref{dh}, where $\alpha_k$ and $a_k$ are the angle and length of rotation from the $Z_{k-1}$ axis to the $Z_k$ axis along the $X_{k-1}$ axis, respectively. $\theta_k$ and $d_k$ are the angle and length of rotation from the $X_{k-1}$ axis to the $X_{k}$ axis along the $Z_k$ axis, respectively.

\begin{table*}[htbp]
\caption{D-H parameter of TIAGo robot arm.}
\label{dh}
\begin{center}
\setlength{\tabcolsep}{7mm}
\renewcommand\arraystretch{1.5}
\begin{tabular}{cccccc}
\hline
Joint Index $k$ & $\alpha_k/rad$  & $a_k/mm$ & $d_k/mm$ & $\theta^l_k/rad$ & $\theta^u_k/rad$\\ \hline
1               & 0               & 0.15505  & -0.151   & 0                & 2.75\\
2               & $\frac{\pi}{2}$ & 0.125    & -0.0165  & -1.57            & 1.09\\
3               & $-\frac{\pi}{2}$& 0        & -0.0895  & -3.53            & 1.57\\
4               & $\frac{\pi}{2}$ & 0.02     & -0.027   & -0.39            & 2.36\\
5               & $-\frac{\pi}{2}$& 0.02     & 0.162    & -2.09            & 2.09\\
6               & $\frac{\pi}{2}$ & 0        & 0        & -1.41            & 1.41\\
7               & $-\frac{\pi}{2}$& 0        & 0        & -2.09            & 2.09\\ \hline
\end{tabular}
\end{center}
\end{table*}

The transformation matrix between the $(k-1)^{th}$ joint and the $k^{th}$ joint of the TIAGo robot arm is shown in \eqref{joint_transform}, where $C_\theta$ and $S_\theta$ donate $\cos{\theta}$ and $\sin{\theta}$, respectively:

\begin{equation}
\label{joint_transform}
    T(\theta_k)^{k-1}_k = 
    \begin{bmatrix}
        C_{\theta_k} & -S_{\theta_k}C_{\alpha_k} & S_{\theta_k}S_{\alpha_k}  & a_kC_{\theta_k} \\
        S_{\theta_k} & C_{\theta_k}C_{\alpha_k}  & -C_{\theta_k}S_{\alpha_k} & a_kS_{\theta_k} \\
        0         & S_{\alpha_k}           & C_{\alpha_k}           & d_k          \\
        0         & 0                   & 0                   & 1            \\
    \end{bmatrix}
\end{equation}

Substituting the parameter in Tab.~\ref{dh}, $T(\theta_1)^{0}_1$, $T(\theta_2)^{1}_2$, ..., $T(\theta_7)^{6}_7$ can be found sequentially, and multiplying them together gives: 

\begin{equation}
\label{fk}
\begin{aligned}
    \prod_{k=1}^7 T(\theta_k)^{k-1}_k &= 
    \begin{bmatrix}
        n_x & o_x & a_x  & p_x \\
        n_y & o_y & a_y  & p_y \\
        n_z & o_z & a_z  & p_z \\
        0   & 0   & 0    & 1   \\
    \end{bmatrix}
    =
    \begin{bmatrix}
        R & P \\
        0 & 1 \\
    \end{bmatrix} \\
    &= M(\theta_1,\theta_2, ..., \theta_7),
\end{aligned}
\end{equation}
where $R$ and $P$ are the pose matrix and position matrix of the end-effector, respectively. $M(\theta_1, \theta_2, ..., \theta_7)$ is the position\&pose matrix of the end-effector, which is the forward kinematic model of the robot.

\subsection{Inverse Kinematic Modeling of Robot Arm}

The robot drives each joint to rotate in order to complete the body motion, so if the end-effector is required to exhibit the desired poses the corresponding angular information of each joint must be obtained by inverse kinematics solution. Since the motion of the robot arm joints is physically constrained, and the degrees of freedom are redundant, this paper uses a particle swarm algorithm to solve the inverse kinematics of the robot arm. First, the seven joint variables $\theta_1$, $\theta_2$, ..., $\theta_7$ of TIAGo robot are transformed into 7-dim position coordinates $P_i(P_{i1}, P_{i2}, ..., P_{i7})$ of the particles. Then the fitness function guides the search direction of the algorithm. Finally the optimal solution of the inverse kinematics problem is found in the solution space under the constraints.

The particle swarm algorithm generates a number of feasible solutions (called particles) randomly in the solution space, uses the fitness function to judge the particles, and lets the particles follow the optimal ones to carry out the motion to find the optimal solution. The optimal particles include the historical optimal particle $pBest$ and the global optimal particle $gBest$. Each particle iteratively updates its velocity $v$ and position $x$ with $pBest$ and $gBest$ as references to explore the solution space:

\begin{equation}
\label{liziqun_speed}
\begin{aligned}
    v(t+1) = W \cdot v(t) &+ C_1\cdot rand() \cdot [pBest(t)-x(t)] \\
    &+ C_2 \cdot rand() \cdot [gBest(t)-x(t)],
\end{aligned}
\end{equation}

\begin{equation}
\label{liziqun_position}
    x(t+1) = x(t) + v(t+1),
\end{equation}
where $W\in [0, 1]$ is the inertia weight, reflecting the effect of the original velocity on the subsequent motion; $C_1$ and $C_2$ are the learning factors, indicating the ability of the particle to utilize its own experience and the ability to absorb the experience of other particles, respectively.

From the current seven joint variables $\theta_1$, $\theta_2$, ..., $\theta_7$, the current position\&pose matrix $M_C$, pose matrix $R_C$ and position matrix $P_C$ of the robot arm end-effector can be calculated by \eqref{fk}. In addition, the desired position\&pose matrix $M_O$, pose matrix $R_O$ and position matrix $P_O$ are given. 

Define the position error $E_P$ as the 2-norm of the difference of the position matrix, i.e.

\begin{equation}
\label{ep}
    E_P = \Vert P_C - P_O \Vert_2
\end{equation}

The current pose matrix $R_C$ and the desired pose matrix $R_O$ are transformed into quaternions $(x_c, y_c, z_c, w_c)$ and $(x_o, y_o, z_o, w_o)$. Define the pose error $E_R$ as

\begin{equation}
\label{er}
    E_R = 2\arccos{(x_o \cdot x_c + y_o \cdot y_c + z_o \cdot z_c + w_o \cdot w_c)}
\end{equation}

Since the Dof of the TIAGo arm are redundant, there exists an infinite set of inverse kinematic solutions for a particular position\&pose in its action space. In order to obtain a unique solution that conforms to the constraint, this paper adds additional conditions with the help of the optimal flexibility rule. For the 7 degrees of freedom TIAGo robot arm, the optimal flexibility is defined as 

\begin{equation}
\label{ofr}
    \min{\{\sum_{k=1}^7[\omega_k(\theta_k(j)-\theta_k(j-1))]^2\}},
\end{equation}
where $\theta_k(j)-\theta_k(j-1)$ is the difference between the current angle and the previous angle of the joint $\theta_k$. $\omega_k$ is the weighting factor, following the principle of "more movement of the lower arm and less movement of the upper arm" to achieve more stable movement. In this paper, we take $\omega_1 = 1$, $\omega_2=\omega_3=0.5$, $\omega_4=\omega_5=\omega_6=\omega_7=0.1$.

According to the position error $E_P$, pose error $E_R$ and the optimal flexibility rule \eqref{ofr}, the fitness function is constructed as 

\begin{equation}
\label{fitness}
    f = \omega_PE_P + \omega_OE_O + \sum_{k=1}^7[\omega_k(\theta_k(j)-\theta_k(j-1))]^2,
\end{equation}
where $\omega_P$ and $\omega_O$, respectively, are the weighting coefficients of $E_P$ and $E_O$, and can be taken as $\omega_P=rand(0,1)$ and $\omega_O = 1-\omega_P$. The smaller the $f$ of the particle, then the better its quality, i.e., the smaller the difference between the current position\&pose matrix $M_C$ and desired position\&pose matrix $M_O$.

To improve the global convergence performance of the algorithm, this paper lets the inertia weights $W$ and the learning factor $C_1$ and $C_2$ make adaptive adjustments with the number of iterations: 

\begin{equation}
\label{ada_wl}
\begin{aligned}
    &W(t)=(W_s-W_e)(\frac{t}{T})^2 + (W_e-W_s)(\frac{2t}{T}) + W_s \\
    &C_1(t) = (C_{1s}-C_{1e})(\frac{t}{T})^2+(C_{1e}-C_{1s})(\frac{2t}{T}) + C_{1s} \\
    &C_2(t) = (C_{2s}-C_{2e})(\frac{t}{T})^2+(C_{2e}-C_{2s})(\frac{2t}{T}) + C_{2s}
\end{aligned}
\end{equation}
where $T$ is the final number of iterations, $t$ is the current number of iterations. Take $W_s=0.9$ and $W_e=0.4$ to denote the initial and final values of $W(t)$, respectively. As the number of iterations $t$ increases, $W(t)$ will gradually become smaller, then the particle swarm can explore the whole solution space at the beginning of the iteration and quickly locate the local area where the optimal solution is located. At the later stage of exploration, the particle swarm can launch a detailed search for the optimal solution locally. Taking $C_{1s}=1.5$ and $C_{1e}=2.5$ to denote the initial and final values of $C_1(t)$, and taking $C_{2s}=2.5$ and $C_{2e}=1.5$ to denote the initial and final values of $C_2(t)$, respectively, which can prevent the algorithm from falling into local optimum at the beginning and enhance the search accuracy at the end.

In this paper, the improved particle swarm optimization algorithm is used to solve the inverse kinematic problem of the TIAGo robot arm, which is described as shown in the Alg.~\ref{alg1}.

\begin{algorithm}[htb]
\caption{Solution of the inverse kinematic problem.}
\label{alg1}
\begin{algorithmic}[1]
    \REQUIRE Desired position\&pose matrix $M_O$;
    \ENSURE Joint variables $(\theta_1, \theta_2, ..., \theta_7)$ of the robot arm;
    \STATE Randomly initialization of 50 7-dim particles;
    \WHILE{The current iteration number $t$ is smaller than the final iteration number $T$}
        \STATE Calculate the fitness of each particle according to \eqref{fitness};
        \STATE Compare the fitness of each particle with the $pBest$, and take the smaller one as the new $pBest$;
        \STATE Take the smallest of the $pBest$ of all particles as $gBest$;
        \STATE Update the weights and learning factors by \eqref{ada_wl};
        \STATE Update the particles by \eqref{liziqun_speed} and \eqref{liziqun_position};
    \ENDWHILE
    \STATE Select the global optimal solution $gBest$;
    \RETURN The 7 joint variables corresponding to $gBest$.
\end{algorithmic}
\end{algorithm}

\subsection{Geometric and Physical Modeling of Home Environment}
The home environment, as a place where humans live, contains a diverse range of objects. To build the digital twin system here, it is necessary to model the household objects in it. Household objects, compared with robots, do not contain complex joint structures and can be modeled directly using Blender software. The basic shape is constructed first, and then given the corresponding materials to obtain a realistic model display effect.

After obtaining the geometric model of each household object, it is also necessary to place them in the correct position. The open source SLAM (Simultaneous Localization and Mapping) technology allows the robot to explore the environment and obtain a environment map. The geometric model of the home environment can be completed by manually placing the household object models and the room structure model in the digital space according to the map.

In order to simulate the manipulation of household items by robots, physical modeling of them is also indispensable. In the Unity simulation engine, physical properties such as gravity and collision can be easily added to various items through various components.

\subsection{Functional Modeling of Household Objects}

\begin{table*}[htbp]
\caption{The Description of Functional Attributes}
\label{fa}
\begin{center}
\setlength{\tabcolsep}{10mm}
\renewcommand\arraystretch{1.5}
\begin{tabular}{ll}
\hline
Functional Attribute & Describe\\ \hline
Pickable     & These objects can be picked up or put down into receptacles.\\
Moveable     & These are non-static objects that can be moved around the scene. \\
Heatable     & These objects can increase the temperature of other objects.\\
Coolable     & These objects can decrease the temperature of other objects.\\
Receptacle   & Receptacle objects allow other objects to be placed on or in them.\\
Toggleable   & These objects can be toggle on or toggle off. \\
Openable     & These objects can be opened or closed.\\
Sliceable    & These objects can be sliced into smaller pieces.\\
Fillable     & These objects can be filled with liquid.\\ \hline
\end{tabular}
\end{center}
\end{table*}

When interacting with household objects, robots cannot only consider physical properties. For example, from the perspective of a physical, both cups and trash cans can hold liquids, but humans will only consider cups rather than trash cans for drinking water. In order to bring robot behavior closer to that of humans, it is also necessary to describe the functional properties of each household item. The functional properties describe the actions that the robot can apply to the object or the functions that the object itself has.

In \cite{licici}, 22 attributes are proposed to describe the functionality of objects in home environment. In this paper, however, we argue that some of these attributes are only relevant to humans and not to robots to accomplish tasks, such as Sittable and Lying. There are also some attributes that can be combined into one attribute. For example, Puttable, Rotatable and Moveable can be unified and described by Moveable. In this paper, we use a total of nine functional attributes as shown in Tab.~\ref{fa} to describe the functional semantic information of household objects.

\subsection{Connection of Home Service Robot Digital Twin System}

Regarding the data interaction in the digital twin system, this paper focuses on the acquisition and transmission of the real-time status of the physical robot platform during its operation, and the control commands from the digital space to the physical robot. With this bi-directional real-time data interaction, on the one hand, the operational data of the physical robot platform will be displayed in real time through the digital space; on the other hand, the digital space can issue control commands to the physical robot to accomplish home service tasks. Specifically, this paper uses ROS to acquire and manage the various data of the physical robotics platform, and then uses ROS\# to achieve two-way communication between the data in ROS and Unity3D.

\section{Experimental Verification and Analysis}

\begin{figure}[htbp]
    \centerline{\includegraphics[width=1\linewidth]{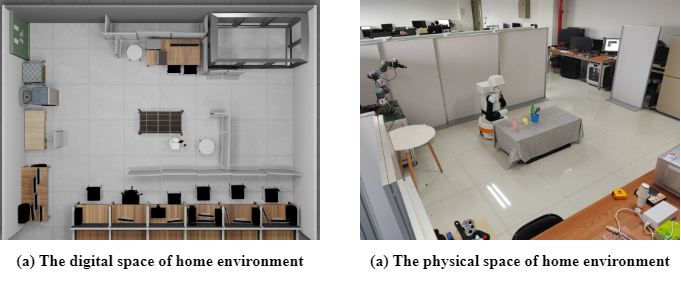}}
    \caption{Simulated home environment built in the laboratory.}
    \label{research_room_3D}
\end{figure}

To validate the proposed approach in this paper, a digital twin system for home service robot is built in laboratory environment as shown in Fig.~\ref{research_room_3D}. The physical space consists of a physical robot platform (TIAGo) and a simulated home environment. The simulated home environment includes household objects such as refrigerator, microwave and dining table which are commonly found in real home environments. In the digital space, the virtual environment uses Unity3D 2021.3.11f1c1 as the development engine. The computer is equipped with a GTX1080 graphics card with 8GB memory, an i7-8700 CPU, and 16GB of RAM.

\subsection{Geometric Model Consistency Verification}

For the geometric model implementation method proposed in this paper, the consistency of the geometric models of the robot platform and the home environment are verified separately.

By measuring the real TIAGo robot, its geometric parameters can be obtained, which allows the calculation of the accuracy of the geometric modeling of the robot platform. In addition, the number of components of the TIAGo robot is known by consulting relevant information, and by comparing the real number with the modeled number, it is possible to indicate whether the robot geometric model has completeness.

Tab.~\ref{jihe_robot} shows the measured data of the physical TIAGo, the modeling data of the digital TIAGo, and the error between the two. All data are in centimeters, except for the number of components. "Height" refers to the overall height of the robot, "Chassis" is the robot's movable chassis, and "Laptop Tray" is the platform behind the head of TIAGo. From the data in Tab.~\ref{jihe_robot}, it can be seen that the modeling error of the geometric model of TIAGo is not large and all the components of TIAGo are modeled, so the geometric model has high accuracy and completeness.

\begin{table*}[htbp]
\caption{Geometric parameters and errors of robot}
\label{jihe_robot}
\begin{center}
\renewcommand\arraystretch{1.5}
\begin{tabular}{cccccccc}
\hline
\multirow{2}{*}{Space} &
  \multirow{2}{*}{\begin{tabular}[c]{@{}c@{}}Components\\ Number\end{tabular}} &
  \multirow{2}{*}{Height(cm)} &
  \multicolumn{2}{c}{chassis} &
  \multicolumn{3}{c}{Laptop Tray} \\ \cline{4-8} 
         &    &          & Height(cm)  & diameter(cm) & Height(cm)  & Width(cm)  & Length(cm) \\ \hline
Physical & 89 & 110      & 30      & 54       & 60      & 28     & 33     \\
Digital  & 89 & 110.0998 & 30.0384 & 53.172   & 60.4548 & 28.476 & 33.264 \\
Error    & 0  & 0.998    & 0.384   & 0.828    & 0.4548  & 0.476  & 0.264  \\ \hline
\end{tabular}
\end{center}
\end{table*}

The geometric model of the home environment is built based on the map acquired by the physical robot performing open source SLAM algorithm. The geometric model of the home environment is subject to errors with the real environment because the map construction generates errors. The coordinates of a vertex of some household objects in the physical space and digital space are measured separately (with the upper left corner of the map as the coordinate origin in centimeters).Since the height coordinates of the object are affected by gravity, only the two-dimensional coordinates of object on the map plane are of interest. The consistency of the geometric model of the home environment is verified by calculating the difference between the coordinates of the two.

\begin{table*}[htbp]
\caption{Geometric parameters and errors of home environment}
\label{jihe_jia}
\begin{center}
\renewcommand\arraystretch{1.5}
\begin{tabular}{cccccccc}
\hline
Space    & Fridge                  & Table1                  & Table2                        & Desk                  & Microwave                  & Television                  \\ \hline
Physical & (107 ,348)              & (412, 157)              & (334, 347)                    & (493, 213)            & (405, 163)                 & (427, 152)                  \\
Digital  & (105.423, 348.525)      & (414.205, 156.423)      & (333.012, 345.423)            & (491.432, 212.433)    & (406.429, 162.956)         & (425.912, 153.422)          \\
Error    & 1.662                   & 0.612                   & 1.861                         & 1.667                 & 1.430                      & 1.790                       \\ \hline
\end{tabular}
\end{center}
\end{table*}

The coordinates of some objects in physical and digital space obtained from the measurements are shown in Tab.~\ref{jihe_jia}, and all data are in centimeters.The error is the Euclidean distance between two coordinates. The data in Tab.~\ref{jihe_jia} shows that the modeling error of the geometric model of the home environment is small, and therefore the geometric model of the home environment is geometrically consistent.

\subsection{Physical Model Validity Verification}

Before connecting the digital space to the physical space, the validity of the physical model designed in this paper is verified by simulating the robot's inverse kinematics in the digital space. This experiment realizes the trajectory planning of the robot arm in the joint space according to the inverse kinematics model by the fifth polynomial interpolation method.

Tasks performed by robots in the home environment, such as delivering objects, require robots with basic grasping capabilities. Therefore, the validity of the physical model is verified by using the example of a robot grasping a water cup.

\begin{figure}[htbp]
    \centerline{\includegraphics[width=1\linewidth]{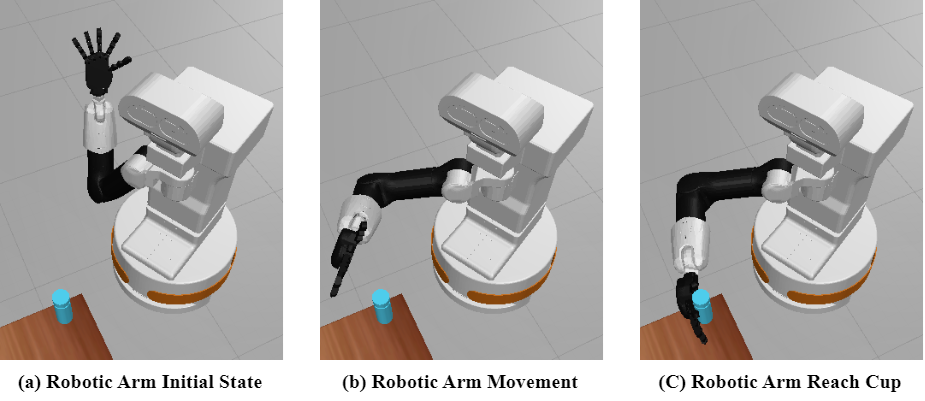}}
    \caption{The process of TIAGo robot hand approaching the water cup.}
    \label{TIAGO_APPROACH}
\end{figure}

To ensure the stability of grasping, the action of TIAGo robot is divided into two stages, approaching and grasping. Fig.~\ref{TIAGO_APPROACH} shows the process of robot approaching the water cup. The robot drives the robot hand to gradually approach the water cup through the position\&pose adjustment of the robot arm until it is close to the water cup.

\begin{figure}[htbp]
    \centerline{\includegraphics[width=0.7\linewidth]{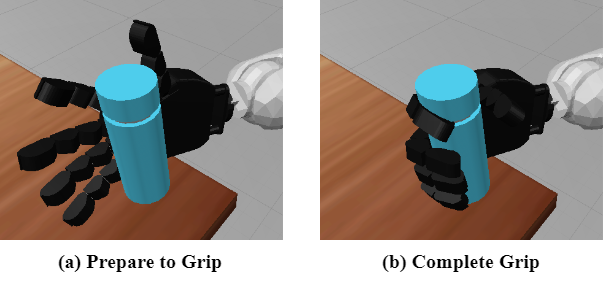}}
    \caption{The process of TIAGo robot hand grasping the water cup.}
    \label{TIAGO_WO}
\end{figure}

After approaching the water cup, the robot hand needs to perform a grasp action, as shown in Fig.~\ref{TIAGO_WO}.

From the experimental results, it can be seen that the physical model designed in this paper enables the TIAGo robot to complete the water cup grasping task smoothly, and the whole grasping process does not have any accidental collision with obstacles and other situations, which meets the expected requirements.

\subsection{Virtual Reality Consistency Verification}

To further validate the method proposed in this paper, the digital twin system is formed by connecting the physical space with the digital space. Since the effective execution of all home service tasks requires robot movement, the virtual-real consistency is verified from the operation of movement command in the digital twin system.

\begin{figure}[htbp]
    \centerline{\includegraphics[width=1\linewidth]{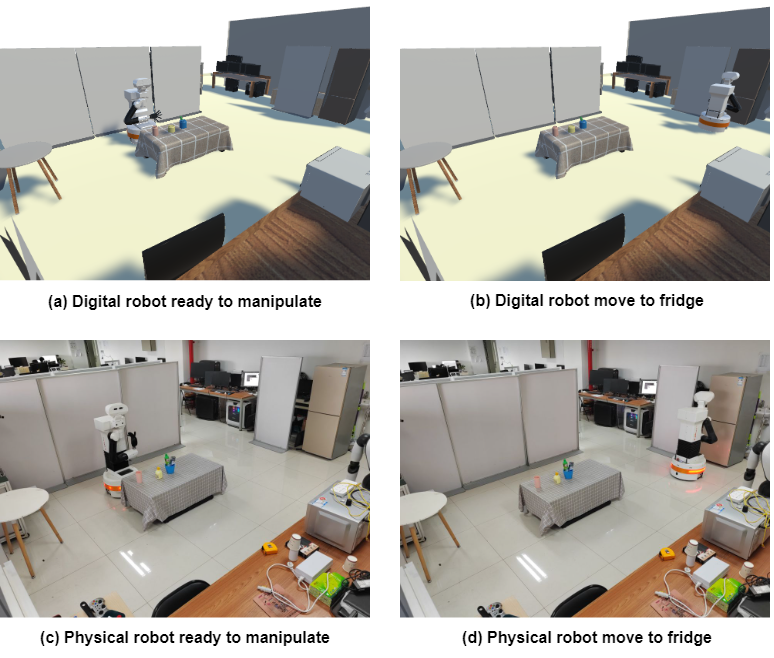}}
    \caption{Operational effect of the digital twin system.}
    \label{TIAGO_tongbu}
\end{figure}

Firstly, the AMCL (Adaptive Monte Carlo Localization) module that comes with the physical TIAGo robot is used to obtain its initial position\&pose in the physical space, and this position\&pose in used to initialize the position\&pose of the digital robot. Secondly, a movement command is sent to the robot in digital space, which will move and display the real-time status of the robot and the home environment in real time. The movements of the physical robot are synchronized in the digital space, and the effect is shown in Fig.~\ref{TIAGO_tongbu}.

As can be seen from Fig.~\ref{TIAGO_tongbu}, the digital space and the physical space can achieve the same execution effect when performing the same task, and the digital space can be synchronized with the physical space in real time. This proves that the digital twin system established in this paper has virtual-reality consistency.

\section{Conclusion}

In order to meet the practical needs of home service robots for complex homework, this paper proposes the motion simulation-based digital twin system for home service robots and its implementation method. This system integrates geometric, physical and functional models for achieving accurate mapping of the home service robot and its working environment. For the geometric model construction of the robot platform, a Unity3D-oriented URDF file parser is designed to automatically construct the 3D model of the robot. For the physical model, the 7 Dof robot arm of the TIAGo is used as an example to model its kinematics, and the particle swarm optimization algorithm is improved for solving the inverse kinematics problem. In addition, to enhance the realism of home environment simulation, a functional model of household objects is proposed, and functional semantic information of household items is described using functional attributes. Finally, the accuracy and practicality of the method and system proposed in this paper are demonstrated through geometric model consistency verification, physical model validity verification and virtual-reality consistency verification, which provide a feasible new approach to the problem of completing complex tasks and other aspects of home service robots.

This study provides a concrete implementation of a digital twin system for home service robot, but there are still missing areas that need a lot of research, such as automatic addition and location update of household items, implementation of more atomic actions, aand the service task planning performed in the digital space. Therefore, subsequent work will further refine the system, such as employing ontology knowledge base to manage functional models and combining intelligent algorithms to implement service task planning for specific research.

\vspace{12pt}
\end{document}